\documentclass[runningheads]{llncs}
\usepackage[T1]{fontenc}
\usepackage{booktabs}
\usepackage{subcaption}
\usepackage{amsmath}
\usepackage{graphicx,verbatim}
\usepackage{orcidlink}
\begin{document}
\title{Weakly-supervised Kidney Tumor Classification from CT Scans with Multi-Instance Learning and Anatomical Filtering}
%

\author{Joonas Ariva\inst{1,2}\orcidlink{0009-0003-7171-7359} \and Dmytro Fishman\inst{1,2,3}\orcidlink{0000-0002-4644-8893}} 
\authorrunning{Ariva et al.}

\institute{University of Tartu, Tartu, Estonia \\
\email{\{joonas.ariva, dmytro.fishman\}@ut.ee} \and
STACC, Tartu, Estonia
 \and
Better Medicine, Tartu, Estonia}
  
\maketitle              
\begin{abstract}
Deep learning models for CT scan analysis are often limited by the scarcity of precise pixel-level annotations, which require significant radiologist effort to produce. Training on scan-level labels alone reduces annotation requirements but introduces challenges: low supervision ratios and large input volumes make models prone to overfitting and shortcut learning. In this work, we investigate two complementary methods to address these challenges: multi-instance learning (MIL) and anatomical filtering. MIL divides CT volumes into 2D slice instances, enabling efficient 
2D architectures with ImageNet pretraining rather than computationally demanding 3D models. Anatomical filtering uses Compass, our self-supervised body part regression model, to crop scans to pathology-relevant subregions without requiring segmentation masks. We evaluate two MIL frameworks — Attention-based MIL (ABMIL) and FocusMIL — on kidney tumor classification 
across one internal dataset (TUH) and two external datasets (KiTS23 and TCGA-KiRC). Our best models achieve F1 = 0.83 on the internal test set using only scan-level labels. We further show that anatomical filtering with the Compass model is 
critical for the out-of-distribution generalization of embedding-based ABMIL, while instance-based FocusMIL demonstrates greater inherent robustness to distribution shift. While evaluated on kidney tumors, we consider this a proof-of-concept for a broader weakly supervised CT classification pipeline applicable to other organs and pathologies.


\end{abstract}
\section{Introduction}

CT imaging plays a central role in the diagnosis of kidney tumors, as well as a wide range of other pathologies, yet the growing volume of scans places an increasing burden on radiologists \cite{lantsman2022trend,radcensus}. Deep learning has emerged as a promising tool to assist in this process with substantial progress being made across a range of tasks including classification, detection and segmentation of pathologies \cite{zhou2021review,ahmad2025deep,gao2025medical}. However, development of such models is often bottlenecked by the scarcity of available data annotations, as these can be expensive to create, especially more precise, pixel-level labels \cite{ahmad2025deep,gao2025medical}. Such annotations must be produced by trained radiologists, who are already occupied with their clinical responsibilities, making the process both time-consuming and expensive.

In light of the data labeling problem, classification models offer a compelling trade-off: they trade precise pathology localisation for considerably reduced labeling requirements. For many clinical scenarios — such as triaging patients or determining whether further examination is warranted — a binary prediction signaling the presence of pathology is sufficient \cite{chilamkurthy2018deep,baltruschat2021smart}. Classification models can provide this signal while requiring only scan-level labels. Such labels can be derived from the accompanying radiology reports, which are routinely generated during clinical practice. Using only scan-level labels reduces the data annotation burden and enables training on larger datasets, which lack the pixel-level labels - a paradigm commonly referred to as weak supervision.

However, scan-level labels are less information-rich than segmentation labels, resulting in a lower supervision ratio. In the segmentation task, the supervision ratio between label and image data is 1:1, meaning that for each pixel (or voxel) there is a label signifying the class of said pixel (either any of the classes or background). In binary classification, this ratio is 1:N, where N is the number of pixels in the image. In CT scans, the voxel count can go up to hundreds of millions. This low supervision ratio makes classification models prone to overfitting to patterns that correlate with the training label, but do not reflect the underlying pathology. Furthermore, the large size of CT scans poses practical challenges for model training, as storing the full volume along with intermediate 
activations and gradients often exceeds available GPU memory.

One way to enhance this supervision ratio is to crop the CT scans to pathology-relevant regions. However, accurate cropping typically requires additional annotations such as bounding boxes or segmentation masks, reintroducing the pixel-level annotation burden that weak supervision aims to avoid. Body part regression (BPR) offers a potential solution: BPR model learns to map CT slices to a continuous anatomical coordinate system in a self-supervised manner, requiring no additional annotations \cite{yan2018unsupervised}. To the best of our knowledge, BPR has not previously been applied to downstream classification tasks. The supervision ratio can also be improved through multi-instance learning (MIL), in which 3D volumes are divided into smaller instances, such as 2D slices, thereby decreasing the instance input dimensionality. MIL has already been applied to CT imaging across a range of pathologies, including lung cancer, COVID-19, and abdominal tumors \cite{han2020accurate,zhao2025lung,cen2025computed,hb2024multi}. However, a common limitation of these approaches is their reliance on organ- or lesion-segmentation masks to define the region of interest prior to training. 

Motivated by these challenges, we propose to combine two complementary methods to improve the training efficiency and classification performance of CT scan models: an anatomical filtering module and multi-instance learning. First, we develop Compass, a self-supervised BPR model that filters CT scans to pathology-relevant regions without requiring segmentation masks, improving the supervision ratio while reducing computational cost. Second, we apply MIL to divide CT volumes into 2D slice instances, enabling the use of efficient 2D architectures with ImageNet pretraining rather than computationally demanding 3D models. As a secondary benefit, reduced input dimensionality may reduce the risk of overfitting to spurious scan-level features.

In summary, the contributions of this work are as follows. We evaluate the performance of two MIL frameworks on the task of kidney tumor classification from CT scans and compare them to a standard whole-scan classification baseline. Furthermore, we combine these models with the Compass model and show that initializing these models from Compass pretrained weights and applying the Compass filtering strategy further improves the performance. To the best of our knowledge, no prior work has applied MIL to kidney tumor classification. While we evaluate our method on kidney tumor classification, we consider it a proof-of-concept for a broader approach that, with future work, could generalize to other organs and pathologies.  

\section{Background}
This section provides background on the two key methodological pillars of our work: multi-instance learning and body part regression.
\subsection{Multi-instance learning}
Training deep learning models directly on whole CT volumes is challenging, as the memory required typically exceeds available GPU capacity. MIL addresses this problem by dividing each image into smaller instances and training the model on instances rather than whole images, reducing the input dimensionality per instance. Instances from the same image are grouped into a collection called a bag. Under the standard MIL assumption, a bag is considered positive if one or more of its instances are positive, and negative otherwise \cite{dietterich1997solving}. 

A typical MIL model has three parts: feature extraction from instances, scoring of the instance features and aggregation into a single output value. Each image patch of a bag is encoded into an embedding vector. Then the instance vectors are aggregated and scored to output the final predicted label for the bag. Aggregation of instances allows for supervising the model on bag-level labels even when instance-level labels are not known. In the setting of CT, instances are typically either 2D slices, 2D patches or 3D patches from a single scan, all sharing a scan-level label.

\begin{figure}[h]
    \centering
    \includegraphics[width=1\columnwidth]{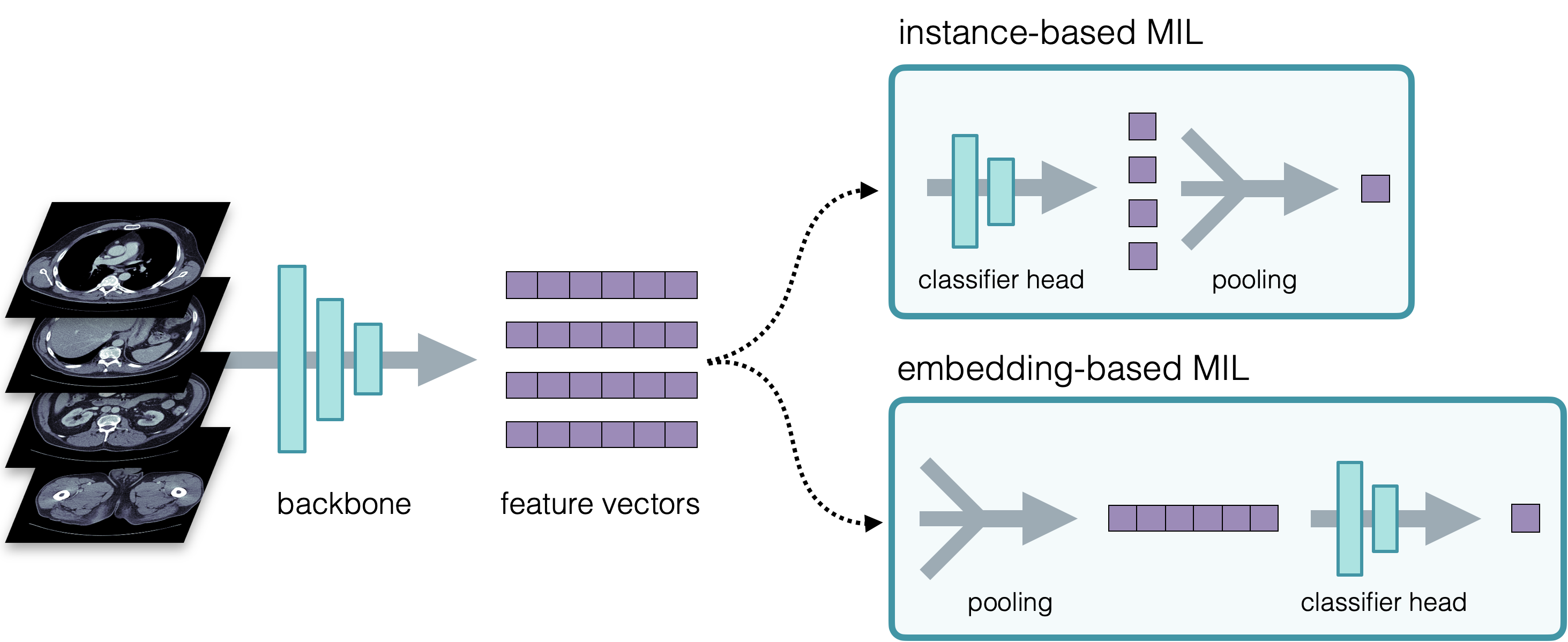}
    \caption{Multi-Instance Learning model designs visualized with example four CT slices. Instance and embedding-based MIL differ by the order of classifier and pooling operations. The backbone encodes each slice into a feature vector. In instance-based MIL models each feature vector is scored with a light-weight classifier and predictions are pooled together to output the bag-level prediction. In our experiments, we use FocusMIL, which uses max pooling operator. In embedding-based MIL models, the feature vectors are pooled into a bag-level feature vector, which is then scored with the classifier for final prediction. In our experiments, we use ABMIL, which uses attention-based pooling.}
    \label{fig:mil}
\end{figure}

MIL model designs differ mainly how instance scoring and aggregation are performed. The designs can be categorized into instance-based and embedding-based approaches (see Figure \ref{fig:mil}) \cite{ilse2018attention}. Instance-based approaches predict a score for each instance vector and pool individual scores together into the final prediction. Common choices for the pooling operator are mean or max pooling. Embedding-based approaches aggregate instance vectors to produce a bag-level vector, which is fed through a final light-weight classifier. Popular pooling here is attention-based pooling introduced by Ilse et al., where an additional attention head predicts attention scores for each vector, which are used as weights to aggregate together the bag-level vector \cite{ilse2018attention}. 

Wang et al. \cite{wang2018revisiting} argue for embedding-based approaches, noting that the performance of instance-based MIL models is dependent on instance-level predictions, for which no ground truth exists - making instance-level prediction itself a weakly supervised problem. Embedding-based approaches avoid this issue by focusing only on the bag-level classification, making it a fully supervised task. On the other hand, Raff and Holt show that many popular embedding-based methods (MI-Net, ABMIL, TransMIL) allow the final classifier to base its prediction on all instances jointly without any constraints \cite{raff2023reproducibility}. This permits a bag to be classified based on the absence of a feature rather than its presence, which directly violates the MIL assumption. Practically, it means that the model might learn to classify bags on causally unrelated features and fail to generalize to out-of-distribution data. These arguments motivate our choice of training and comparing both embedding-based and instance-based MIL models on the task of kidney tumor classification and also evaluating the trained models on out-of-distribution datasets.

\subsection{MIL and weak supervision in CT imaging}

Prior work applying MIL to CT modality has had a large focus on lung-related pathologies. MIL has been used to classify COVID-19 \cite{han2020accurate,chikontwe2021dual,he2021synergistic}, chronic obstructive pulmonary disease \cite{xu2020dct,frade2022multiple,xue2023ct} and lung cancer \cite{zhao2025lung,frade2022multiple} from chest CT scans. Lung pathologies are well-suited to MIL as diseased regions tend to be spatially localized,producing a discriminative signal at the instance level.

More recently, MIL has been extended to abdominal CT imaging, with applications to hepatocellular carcinoma staging and prognosis \cite{cen2025computed,chang2024revit}, colorectal cancer lymph node metastasis \cite{xie2023predicting}, and ovarian tumor classification \cite{hb2024multi}. To the best of our knowledge, however, MIL has not previously been applied to kidney tumor classification.

A common thread throughout the MIL research on CT scans - both thoracic and abdominal - is the reliance on organ or lesion segmentation masks as a preprocessing step to define the region of interest prior to training. In the context of lung it involves typically the use of segmentation models to segment lungs and training only on that area \cite{han2020accurate,xu2020dct,chikontwe2021dual,frade2022multiple,zhao2025lung,xue2023ct,he2021synergistic}. In abdominal regions, equivalent organ-level or lesion-level masks are used to crop the input before applying MIL \cite{cen2025computed,chang2024revit,xie2023predicting,hb2024multi}.
Considering the whole training pipeline, these methods are therefore not truly weakly supervised as they depend on supervised segmentation models trained on pixel-level annotations. This introduces additional annotation requirements and limits generalizability to new anatomical settings. In this work, we adopt a  weaker supervision regime — requiring only scan-level labels for training and a small set of slice-level boundary annotations for anatomical calibration, without pixel-level segmentation masks at any stage. This commitment to weak supervision, while more challenging, enables a more scalable pipeline that could in principle be applied to any organ or pathology without additional annotation effort.

\subsection{Body part regression}
BPR offers a segmentation-free approach to anatomical localization in CT scans, making it a natural candidate for defining pathology-relevant subregions without pixel-level annotations \cite{yan2018unsupervised}. A BPR model predicts a continuous value for each slice, which reflects the position of the slice inside the body. Successfully trained BPR model attributes similar scores to slices depicting the same body regions in different scans. The task of BPR is well-suited to self-supervised training as each scan inherently has the correct ordering of the slices, which can be used as supervisory signal, eliminating the need for external labels for training. Beyond predicting the order of axial slices, later works have developed models, which predict relative 3D coordinates from 3D patches of CT scans \cite{lei2021contrastive} and even 3D coordinates for each scan voxel \cite{goncharov2024anatomical}. Despite this body of work, to the best of our knowledge, no prior work has examined the effect of BPR on downstream pathology classification tasks. In this work, we develop our own self-supervised BPR model called Compass and study its effect on reducing input data size for a kidney tumor classification task.

\section{Methods}
\begin{figure}[h]
    \centering
    \includegraphics[width=1\columnwidth]{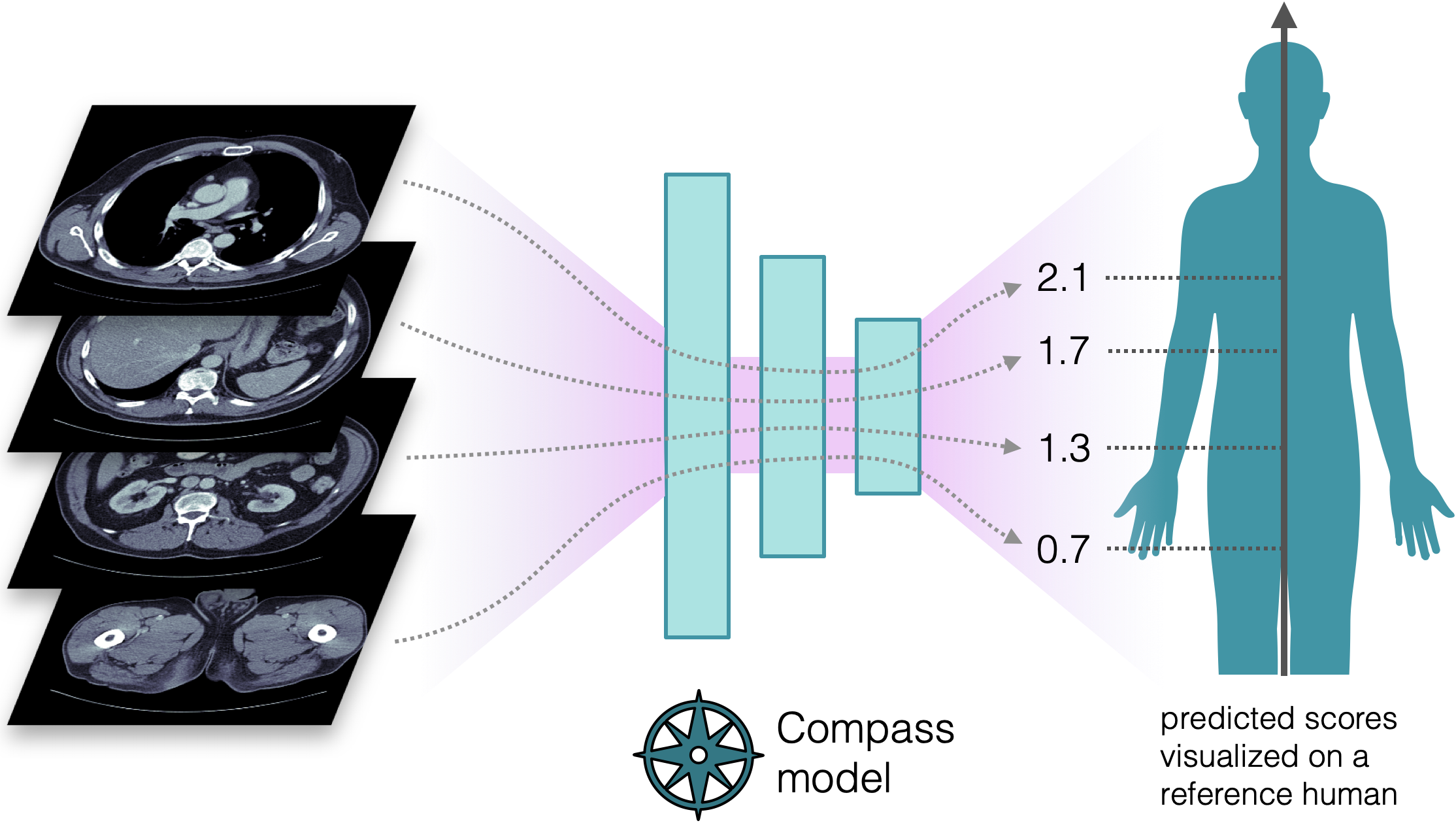}
    \caption{Compass model inference for distinctive key slices is visualized with example four CT slices. Each slice is mapped to a Compass score, which shows the position of the slice on a reference human. The relative distances between Compass scores correlate with respective relative slice distances in the scan.}
    \label{fig:comp}
\end{figure}

Our classification model training happens in two phases. At first, we train a Compass model (visualized in  Figure \ref{fig:comp}), calibrate it for the kidney region, and use the calibrated model to filter the dataset. Then, we train a MIL model on the reduced dataset on the task of kidney cancer classification.
\subsection{Compass model}

Let $x_i$ denote the $i$-th axial slice sampled from a CT scan. Then the predicted pairwise distance between slices $x_i$ and $x_j$ from the same scan is given by $d_{i,j}^{pred} = f(x_i) - f(x_j)$, where $f(\cdot)$ denotes the Compass model, which maps CT scan slice to a scalar value (a Compass score) and is implemented as a ResNet18 encoder \cite{he2016deep} with a single output neuron, initialized with ImageNet-pretrained weights \cite{deng2009imagenet}. True distance between slices is $d_{i,j}^{true}=  n \cdot \lambda_{spacing} \cdot (i - j) $, where $\lambda_{spacing}$ denotes axial slice thickness of the scan in millimeters and $n$ is the sampling stride between consecutive slices in the scan ($n =1$, when every slice is used and $n > 1$ under memory-constrained subsampling). The Compass loss function is constructed by feeding the scan with $N$ slices through the model and calculating L1 loss between predicted and true distances: 

\begin{equation} \label{depth_loss}
L_{compass} = \frac{1}{N(N-1)}\sum_{\substack{i,j=1\\ i\neq j}}^N \left |d_{i,j}^{pred} - \lambda_{scale}\cdot d_{i,j}^{true}\right |,
\end{equation}
where $\lambda_{scale}$ is a fixed scaling constant. Minimizing $L_{compass}$ encourages the model to assign scores proportional to true anatomical distances, such that similar anatomical regions receive consistent scores across different scans.

After the Compass model is trained, it is calibrated on a small set of annotated scans from the training dataset  — in our experiments, 20 scans (10 with kidney tumors and 10 without), each requiring only two slice-level annotations. For each scan in the calibration set, two slices $x_{top}$ and $x_{bottom}$ are extracted: the topmost and bottommost slices that contain the kidney or its tumor. Compass scores $f(x_{top})$ and $f(x_{bottom})$ are subsequently predicted for each calibration scan. The 5th and 95th percentiles of the bottom and top slice scores respectively define the final calibration thresholds $c_{min}$ and $c_{max}$. Compass scores are predicted for every slice $x$ in the dataset and only slices with  $c_{min} \leq f(x) \leq c_{max}$ are used for the classification model training and testing. 

\subsection{Classification models}

We compare three model architectures: a 3D CNN baseline, an embedding-based MIL model, and an instance-based MIL model. For the embedding-based approach we use Attention-based MIL (ABMIL) \cite{ilse2018attention}, which aggregates instance feature vectors into a bag-level embedding via a lightweight attention layer. For the instance-based approach we use FocusMIL \cite{liu2024correlation}, a max-pooling MIL model designed to address the training challenges of standard max-pooling approaches such as training instance memorization. FocusMIL introduces a variational information bottleneck after the feature extractor, which acts as a regularizer by compressing instance representations with Kullback-Leibler loss to retain only task-relevant information. This helps to prevent the model from overfitting to instance-specific noise. Both MIL models use a ResNet18 backbone pretrained on ImageNet. The 3D baseline processes the full scan volume through a 3D ResNet18 backbone. We experimented with initializing the 3D backbone using MedicalNet weights \cite{chen2019med3d}, pretrained on a large collection of medical imaging segmentation tasks, but found this did not improve over random initialization in our setting. We therefore report results with random initialization only.

\subsection{Datasets}
\textbf{TUH Kidney Tumor Dataset.} For model training, we used scans collected from Tartu University Hospital (TUH). This dataset consist of 363 contrast-enhanced CT scans of malignant kidney tumors and 723 control scans. A held-out test set of 100 cases and 100 controls was reserved for evaluation. The dataset is annotated by TUH radiologists on pixel-level for classes \textit{malignant kidney lesion}, \textit{benign kidney lesion} and \textit{kidney tissue}. Its important to note that these labels were only used for evaluation and Compass model calibration purposes - Compass model was trained without labels and classification models used only scan-level labels.

\textbf{KiTS23 \& TCGA-KiRC Datasets.} To assess generalization to 
out-of-distribution data, we additionally evaluated our models on two public kidney tumor datasets. KiTS23 \cite{heller2023KiTS21} contains 489 contrast-enhanced CT scans with publicly available kidney and tumor segmentation masks. TCGA-KiRC \cite{akin2016tcgaKiRC} focuses on kidney renal clear cell carcinoma and includes a variety of imaging modalities; in this study we use a subset of 180 contrast-enhanced CT scans, annotated in house by a radiologist with the same label classes as the TUH dataset. As both datasets consist exclusively of tumor-positive cases, evaluation on these datasets is limited to sensitivity (recall).

\textbf{Data Processing.} All CT scans were resampled to an isotropic voxel spacing of 2~mm $\times$ 2~mm $\times$ 2~mm to standardize the input across scanners and acquisition protocols while preserving clinically relevant anatomical detail. Axial slices were cropped and padded to 250 $\times$ 250 pixels to remove empty border regions. Soft tissue windowing was applied by clipping Hounsfield unit (HU) values to the range $[-150, 250]$, followed by normalization to $[0, 1]$. For 2D MIL models, each sampled slice was stacked with its two immediate neighbours along the axial axis to form a three-channel input, compatible with the three-channel architecture of ImageNet-pretrained backbones. Training data was augmented with random horizontal flips, 90-degree rotations, mild Gaussian noise ($\sigma = 0.05$), and mild intensity scaling ($\pm 10\%$). 

\subsection{Implementation details}

All models were implemented in PyTorch \cite{paszke2019pytorch} using the MONAI framework \cite{cardoso2022monai} for CT scan processing. Training was performed on 8 AMD MI250X GPUs using distributed data parallel training and automatic mixed precision. Models were trained for 150 epochs using AdamW optimizer with an initial learning rate of 1e-4, weight decay of 5e-4 and cosine annealing scheduler with a linear warmup over first 10\% of epochs. For mini-batch experiments the batch size of five was used; all other configurations used a batch size of one. To address class imbalance in the TUH dataset a weighted random sampler was used in the dataloader. Binary cross-entropy loss was used for classification experiments, with additional Kullback-Leibler divergence loss for variational information bottleneck in FocusMIL, weighted by a coefficient of 0.1. For Compass model training, Compass loss was used as specified in Equation \ref{depth_loss} with $\lambda_{scale}=0.1$ and slice sampling stride $n = 1$. 

\section{Experiments \& Results}
We evaluate our approach in two stages: first assessing the Compass module and then evaluating the full classification pipeline across multiple model configurations.

\subsection{Effects of Compass model filtering}
First, we evaluate Compass model independently to demonstrate its ability to retain relevant information from the scan while filtering out non-relevant slices. We train a self-supervised Compass model on a TUH train set and evaluate it with different calibration set sizes. For this we sample $n$ scans from TUH training set for Compass model calibration and then evaluate its performance on the remaining TUH dataset. We vary $n$ from $2$ to $200$ and sample equally from both classes to assess the effect of calibration size on filtering quality. We calculate two separate metrics to evaluate the effectiveness of the filtering: tumorous slice recall and average scan size reduction. Tumorous slice recall shows what percentage of tumorous slices out of all tumorous slices in the dataset are retained by filtering. This is important as the label of the scan is defined by the presence or absence of tumorous slices. Average scan size reduction measures the mean ratio between the number of slices discarded relative to the total volume size. We also check that no tumorous scans have all their tumor-containing slices removed by the filter, which would incorrectly flip the scan-level label. For each sample size $n$, we repeat the experiment 100 times and report the average metrics. We also report the same metrics on KiTS and KiRC datasets (still with calibration on TUH) to demonstrate its effectiveness on out-of-distribution data.

\begin{figure}[h]
    \centering
    \includegraphics[width=1\columnwidth]{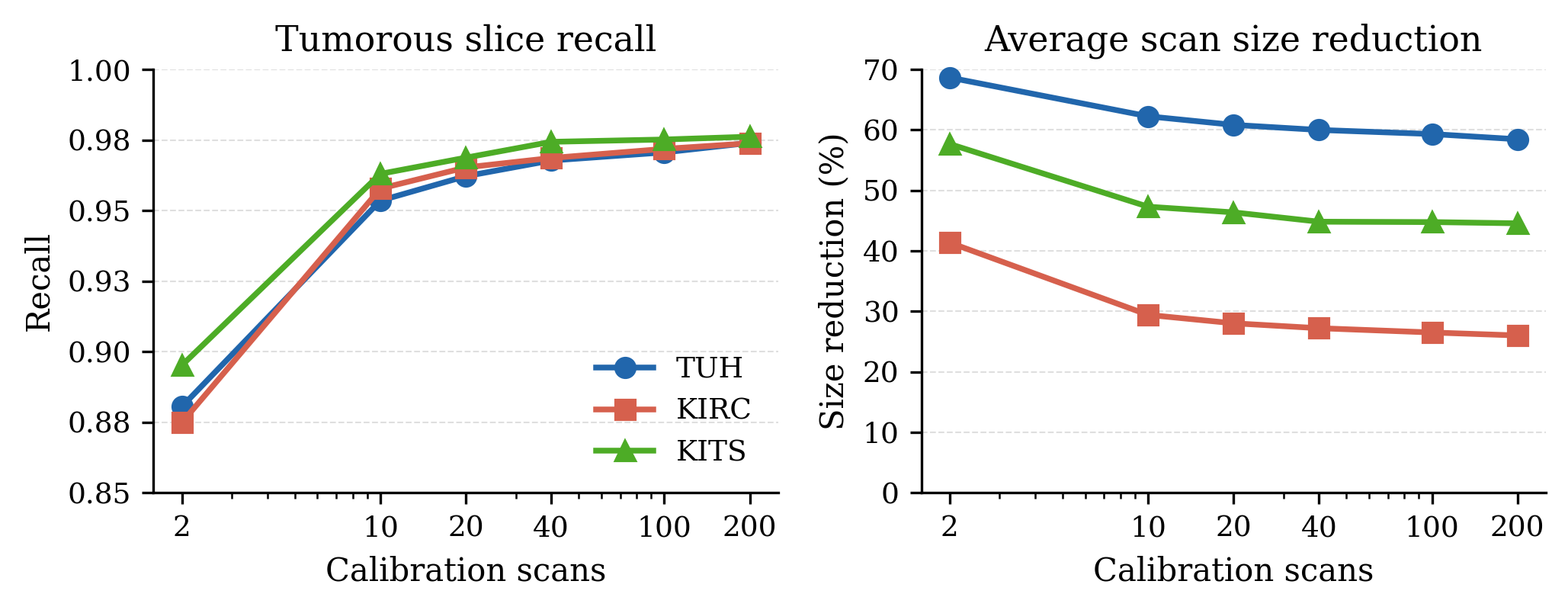}
    \caption{Results of Compass filtering evaluation. X-axes are on log scale to better highlight different calibration set sizes. Compass filtering can reduce average scan size considerably, while keeping the task-relevant slices for model training.}
    \label{fig:comp_reduction}
\end{figure}

\textbf{Results.} Compass filtering retains the vast majority of tumor-containing slices across all datasets, with tumorous slice recall on TUH ranging from 0.87 ($n=2$) to 0.97 ($n=200$), and only ten calibration scans needed to reach a recall of 0.95. Evaluation on out-of-distribution datasets KiTS and KiRC yield similar tumor recalls, with performance plateauing beyond ten calibration samples. Average scan size reduction varies between 68\% ($n=2$) and 58\% ($n=200$) for TUH. Evaluation on KiTS and KiRC datasets yield similar results with slightly smaller scan reductions (KiTS: 57\%-44\%, KiRC: 40\%-26\%), signifying that these datasets are already more kidney-focused and there are fewer non-relevant slices to remove. Overall, the size reductions get smaller as the calibration size increases. This is expected as the Compass filtering thresholds slightly widen with more variety introduced to the calibration set. In no experiment did Compass filtering remove all tumorous slices from any scan, confirming that scan-level labels are never incorrectly flipped by the filtering. Full results are shown in Figure~\ref{fig:comp_reduction}

\subsection{Classification experiments}

We evaluate three model configurations: 3D ResNet18 (baseline), ABMIL, and FocusMIL. For each configuration, we test with and without Compass filtering, and additionally evaluate Compass pretraining for MIL models. As the original FocusMIL uses mini-batch gradient descent, we train both MIL models with and without it to assess its contribution in the CT domain. Modifications are applied cumulatively — each configuration in Table~\ref{tab:main_results} builds on the previous one within its model group, rather than testing each modification in isolation. All models are trained on the TUH training set and evaluated on the TUH test set, KiTS, and KiRC. Each model configuration is trained three times and we report the mean performance at the epoch with the highest F1 score on the TUH test set.

We also experimented with a frozen backbone strategy, training only the MIL aggregation head while keeping the ResNet18 backbone fixed — both with ImageNet-pretrained and Compass-pretrained initializations. However, this consistently yielded poor classification performance, suggesting that fine-tuning the full network is necessary for adapting pretrained features to the CT domain. All reported results therefore use end-to-end fine-tuning.

\textbf{Results.} The 3D baseline model without Compass filtering collapses to predicting the positive class for every scan (recall = 1.0, specificity = 0.0), indicating that training the 3D architecture from scratch is ineffective at this dataset size.  Compass filtering partially alleviates this, yielding F1 = 0.70, though performance remains below the MIL-based approaches. Switching to 2D MIL models improves the classification performance. ABMIL achieves consistently test F1 score of $\approx$ 0.82 even without Compass filtering, demonstrating the benefit of ImageNet-pretrained features and the benefit of a more favorable instance-level supervision ratio. 
However, the external validation reveals a critical limitation: without Compass filtering, ABMIL recall drops to 0.36 and 0.55 on KiTS and KiRC respectively, despite stronger TUH performance. This discrepancy is consistent with the known tendency of embedding-based MIL to exploit the absence of features as a classification signal, leading to non-causal predictions that fail to generalize to out-of-distribution data. Compass filtering largely recovers this generalization gap, improving KiTS and KiRC recall to 0.59 and 0.78 respectively, with minimal impact on TUH performance (F1 = 0.81). 

FocusMIL does not exhibit the same generalization gap without Compass filtering (KiTS recall = 0.52, KiRC recall = 0.67), consistent with instance-based MIL models being less susceptible to non-causal feature exploitation. Compass filtering provides a modest but consistent improvement also for FocusMIL. Adding both Compass-pretrained weights and mini-batch gradient descent yields the strongest FocusMIL result, achieving F1 = 0.83. For ABMIL, adding Compass pretrained weights and mini-batch training achieves F1 = 0.83, matching the best FocusMIL result. However, the relative gain over the ABMIL baseline is smaller, as ABMIL already achieves a stronger baseline performance (F1 = 0.82) compared to FocusMIL (F1 = 0.76). This asymmetry is expected: max-pooling MIL models are generally harder to train than embedding-based approaches, as gradients flow only through the single max-pooled instance per bag, whereas in embedding-based MIL all instances contribute to the gradient. Compass pretraining and mini-batch training both help address this optimization difficulty, explaining the larger relative gain for FocusMIL.  

Overall, instance-based FocusMIL and embedding-based ABMIL reach comparable peak performance on TUH, but differ substantially in out-of-distribution generalization, with Compass filtering being critical for ABMIL but less so for FocusMIL. A consistent gap between in-distribution (TUH) and out-of-distribution (KiTS) recall across all configurations further highlights the generalization challenges of cross-site kidney tumor classification. Full results are given in 
Table~\ref{tab:main_results}.

\begin{table}[]
\centering
\begin{tabular}{@{}llccc|cc@{}}
\toprule
            & \multicolumn{4}{c|}{TUH test} & KiTS   & KiRC   \\ \midrule
Method      & F1  & Recall & Precision & Specificity & \multicolumn{2}{c}{Recall} \\ \midrule
3D baseline & 0.66 $\pm$ 0   & 1       &     0.5      &0&   0.99     &   1     \\
+ Compass   & 0.70 $\pm$0.03   &  0.77      &   0.64      & 0.56 & 0.65       & 0.61       \\ \midrule
ABMIL       & 0.82 $\pm$ 0.02  &   0.79     &   0.86       &0.88 &  0.36      &   0.55     \\
+ Compass   & 0.81 $\pm$ 0.02    &   0.79     &   0.84       &0.85 &  0.59      &    0.78    \\ 
+ pretrain.   & 0.82 $\pm$ 0.01  & 0.83  &   0.82        &0.82 &  0.54     &  0.76      \\
+ mini-batch   & 0.83 $\pm$ 0.01   & 0.83    & 0.83       & 0.84 &  0.5      &    0.69    \\\midrule
FocusMIL    & 0.76 $\pm$ 0.01   &   0.70     &    0.83      &0.85 &   0.52     &   0.67     \\
+ Compass   & 0.77 $\pm$ 0.01   &   0.72    &   0.83       &0.85 &    0.55    &   0.71    \\
+ pretrain.   & 0.80 $\pm$ 0.01   &   0.76    &   0.85      & 0.87 &   0.54     &  0.75      \\
+ mini-batch   & 0.83 $\pm$ 0.02   &  0.77   & 0.89       & 0.91 &   0.5     &  0.68      \\
\bottomrule
\end{tabular}
\caption{Main classification results. Each row cumulatively adds to the configuration above it within each model group. Pretrain. means that classification model is initialized from Compass pretrained weights. For F1 score we also report the standard deviation over three runs. For KiTS and KiRC only recall is reported as these datasets contain only tumorous scans. }
\label{tab:main_results}
\end{table}

\section{Discussion}
The Compass model demonstrates strong anatomical localization performance on kidney CT scans, accurately narrowing volumes to the relevant subregions without discarding diagnostically important details. Notably, this filtering requires minimal supervision - only two slice-level boundary annotations per calibration scan, with as few as 10-20 calibration scans sufficient to generalize across the dataset. The benefits of Compass filtering are twofold: it reduces computational cost by decreasing the input data size, and it improves the supervision ratio by removing slices that contain no task-relevant information. 

Classification experiments show that Compass improves the 3D baseline, preventing it from collapsing to always predict the positive class. At the same time, MIL models remain stable even without Compass filtering, suggesting that processing individual slices rather than full volumes acts as an implicit regularizer in low-data regimes. For MIL, Compass does not substantially improve TUH dataset test performance, suggesting that MIL is capable of identifying relevant training signal even within unfiltered scans. However, Compass filtering plays a critical role in preventing a failure mode of ABMIL: without it, ABMIL generalizes poorly to out-of-distribution datasets. Compass filtering appears to mitigate this by constraining the input to task-relevant regions, reducing the opportunity for the model to learn spurious correlations from irrelevant scan regions. FocusMIL does not exhibit such a failure mode, consistent with the theoretical arguments against embedding-based MIL raised by Raff and Holt \cite{raff2023reproducibility}.

These findings also highlight a broader advantage of MIL over direct 3D classification. The 3D baseline requires Compass filtering to avoid complete prediction collapse, whereas MIL models remain stable even on unfiltered inputs. This suggests that the reduced input dimensionality - processing individual slices rather than full volumes - is itself a regularizing factor, which makes MIL a more robust choice for CT classification in low-data regimes. 

When evaluating models, we noticed that one common error made by models was misclassifying large benign cysts as malignant tumors and predicting such slices as tumorous. This suggests that the task is not simply discriminating between normal and abnormal kidney tissue, but also requires distinguishing malignant abnormalities from benign abnormalities. One potential solution for this would be to include cyst labels in training and turn this classification problem into multi-label classification task. 

A broader challenge for CT classification in weak label regimes stems from the structure of publicly available datasets. Public datasets are typically focused on a single pathology, as exemplified by KiTS \cite{heller2023KiTS21}, TCGA-KiRC \cite{akin2016tcgaKiRC}, LiTS \cite{bilic2023liver}, and PanTS \cite{li2025pants}. Such datasets contain only positive cases, with no control scans from the same acquisition site. This creates conditions for shortcut learning, where models learn to associate scanner-specific artifacts (radiation dose, reconstruction kernel, scanner model) with the positive label rather than the pathology itself. We observed this directly when we divided KiRC and KiTS datasets between testing and training sets: Models appeared to recognise the acquisition style of these datasets, which contained only tumorous scans, resulting in inflated test performance compared to the TUH test set. This challenge can be addressed in several ways. The most straightforward solution is to collect new datasets containing both positive and negative cases from the same acquisition site. Alternatively, medical image harmonization methods could be applied to reduce site-specific style differences across datasets \cite{yadav2025comparative,liu2024learning}. Finally, acquisition style differences could be handled through explicit MIL model design like Interventional Bag Multi-Instance Learning \cite{lin2023interventional}.

Furthermore, another consequence of the absence of control scans in both KiTS and KiRC is that it limits external evaluation to sensitivity only, precluding the computation of specificity or precision on these datasets. This makes it difficult to assess whether strong external recall reflects genuine generalization or simply a bias toward predicting the positive class on out-of-distribution data.

\section{Conclusion}

We presented a weakly supervised method for kidney tumor classification from CT scans, combining an anatomical 
filtering model Compass with multi-instance-learning. Our results demonstrate that Compass model is an effective way of narrowing the CT scan down to the task-specific region without the use of strong supervision. The experiments with classification models combined with the Compass filtering show that weakly-supervised classification is achievable using only scan-level labels, with the best models reaching F1 = 0.83 on the held-out TUH test set. While these results demonstrate the promise of weakly supervised MIL for kidney tumor classification, further work is needed to improve both absolute classification performance and cross-site generalization before  clinical deployment could be considered.
\begin{credits}
\subsubsection{\ackname} We acknowledge the support of the European Union and the Estonian Research Council through project TEM-TA101. We also acknowledge ETAIS (Estonian Scientific Computing Infrastructure) for awarding this project access to the LUMI supercomputer, owned by the EuroHPC Joint Undertaking, hosted by CSC (Finland) and the LUMI consortium through ETAIS, Estonia. Additional computational resources were provided by the High-Performance Computing Cluster at the University of Tartu. We also want to thank Better Medicine and Tartu University Hospital for their support. 

\end{credits}

\bibliographystyle{splncs04}
\bibliography{mybibliography}

\end{document}